\PassOptionsToPackage{table}{xcolor}
\documentclass[]{bytedance_seed}

\usepackage[toc,page,header]{appendix}

\usepackage{minitoc}
\usepackage{amsmath} 
\usepackage{bm} 
\usepackage{booktabs}
\usepackage{graphicx}
\usepackage{makecell}
\usepackage{lipsum}
\usepackage{enumitem}
\usepackage{algorithm}
\usepackage{algpseudocode}
\usepackage{amsmath,amssymb,amsthm}
\usepackage[table]{xcolor}
\usepackage[most]{tcolorbox}
\usepackage{stmaryrd}
\usepackage{todonotes}
\usepackage{amsthm}
\usepackage{parskip} 
\usepackage{wrapfig}
\usepackage{subcaption}
\usepackage{xcolor}
\usepackage{xspace}
\usepackage{marvosym}
\usepackage{amsfonts}       
\usepackage{longtable}      
\usepackage{arydshln}

\newtcolorbox{examplebox}[1]{colback=gray!10!white, colframe=black!75, title=#1, fonttitle=\bfseries, arc=0mm, boxsep=1pt, left=2pt, right=2pt, top=2pt, bottom=2pt}
\usepackage{tabularx}
\usepackage{CJKutf8}

\usepackage{makecell}

\usepackage{pifont}
\newcommand{\cmark}{\ding{51}} 
\newcommand{\xmark}{\ding{55}} 

\title{MOSAIC: Multi-Subject Personalized  Generation via Correspondence-Aware Alignment and Disentanglement}

\author[\spadesuit*]{Dong She}
\author[\spadesuit* \dagger ]{Siming Fu}
\author[\heartsuit*]{Mushui Liu}
\author[\spadesuit*]{Qiaoqiao Jin}
\author[\diamondsuit\spadesuit*]{Hualiang Wang}
\author[\spadesuit]{Mu Liu}
\author[(\text{\Letter})]{\authorbreak Jidong Jiang}

\affiliation[\spadesuit]{ByteDance Fanqie}
\affiliation[\heartsuit]{Zhejiang University}
\affiliation[\diamondsuit]{The Hong Kong University of Science and Technology}

\contribution[*]{Equal contributions}
\contribution[\dagger]{Project lead}
\contribution[\text{\Letter}]{Corresponding authors}

\abstract{

Multi-subject personalized  generation presents unique challenges in maintaining identity fidelity and semantic coherence when synthesizing images conditioned on multiple reference subjects.
Existing methods often suffer from identity blending and attribute leakage due to inadequate modeling of how different subjects should interact within shared representation spaces.
We present \textbf{MOSAIC}, a representation-centric framework that rethinks multi-subject generation through explicit semantic correspondence and orthogonal feature disentanglement.
Our key insight is that multi-subject generation requires precise semantic alignment at the representation level—knowing exactly which regions in the generated image should attend to which parts of each reference.
To enable this, we introduce SemAlign-MS, a meticulously annotated dataset providing fine-grained semantic correspondences between multiple reference subjects and target images, previously unavailable in this domain.
Building on this foundation, we propose the semantic correspondence attention loss to enforce precise point-to-point semantic alignment, ensuring high consistency from each reference to its designated regions. Furthermore, we develop the multi-reference disentanglement loss to push different subjects into orthogonal attention subspaces, preventing feature interference while preserving individual identity characteristics.
Extensive experiments demonstrate that MOSAIC achieves SOTA performance on multiple benchmarks.
Notably, while existing methods typically degrade beyond 3 subjects, MOSAIC maintains high fidelity with \textbf{4+ reference subjects}, opening new possibilities for complex multi-subject synthesis applications.

}

\checkdata[Project Page]{\url{https://bytedance-fanqie-ai.github.io/MOSAIC}}

\begin{document}
\maketitle

\begin{figure*}[!t]
    \centering
    \vspace{-8mm}
    \includegraphics[width=1.0\linewidth]{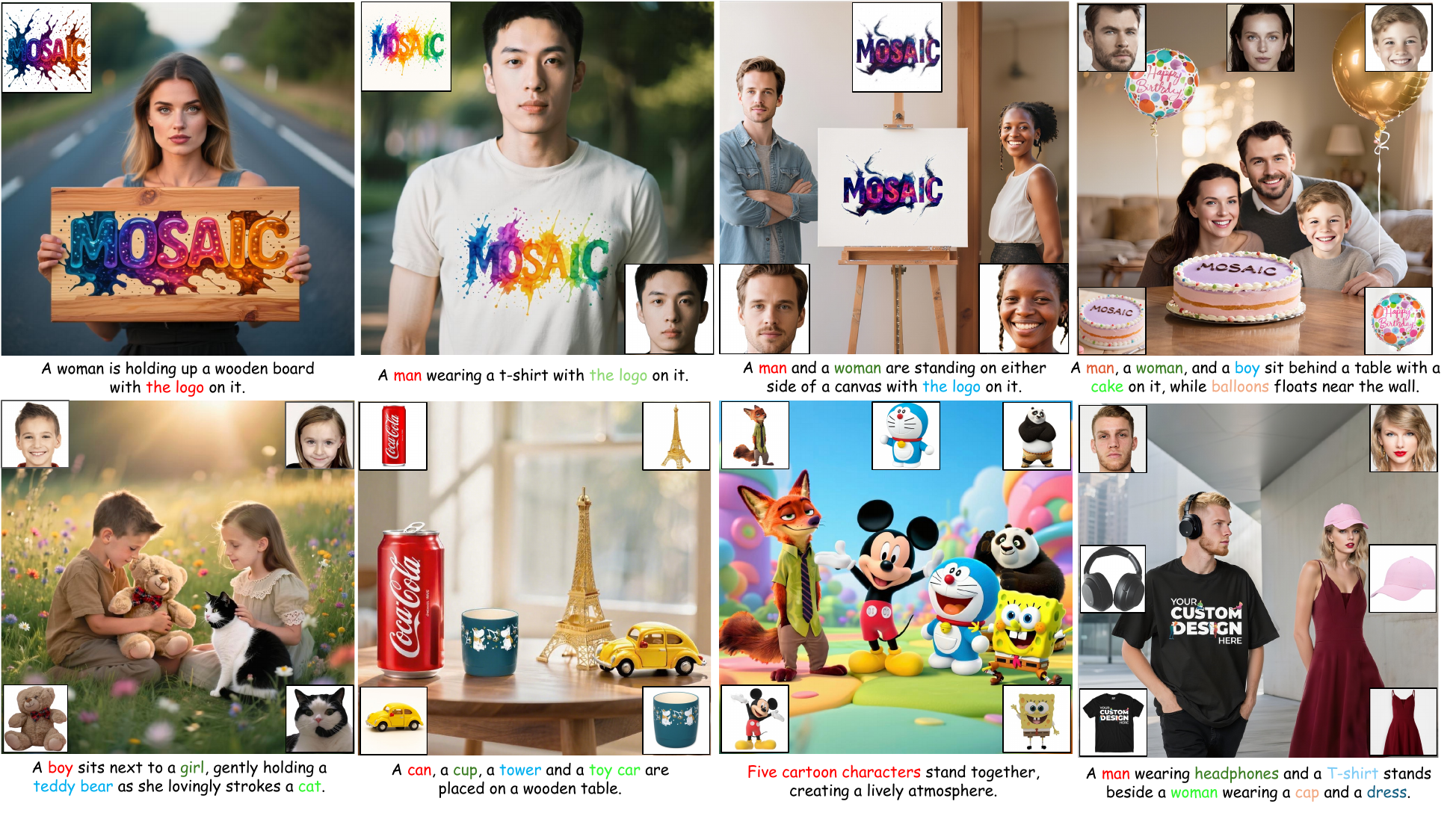}
    \captionof{figure}{Our proposed MOSAIC demonstrates capabilities in single-subject and multi-subject driven generation tasks. } 
    \label{fig:teaser}
\end{figure*}

\section{Introduction}

Multi-subject personalized generation in controllable image synthesis faces significant challenges in maintaining identity consistency while preventing attribute entanglement.
Recent approaches have explored various strategies to address this problem: MS-Diffusion~\cite{wang2024ms} and SSR-Encoder~\cite{zhang2024ssr} incorporate spatial layout guidance into cross-attention layers to bind subjects to dedicated regions, while DreamO~\cite{mou2025dreamo} embeds routing constraints directly into DiT blocks for architectural-level control. Most recently, XVerse~\cite{xverse} introduces token-specific modulation offsets for independent subject representation control.

However, these methods share a critical limitation: they lack explicit optimization of the underlying diffusion representations for both precise multi-subject alignment with target images and effective disentanglement between reference subjects.
This shortcoming becomes increasingly severe as subject count grows, with most methods experiencing significant degradation beyond 3-4 subjects due to compounding feature interference.
To understand why existing approaches fail, we analyze the fundamental requirements for multi-subject generation and identify two key deficiencies.
The first concerns the correspondence problem: without explicit modeling of which specific reference image regions should attend to target latent parts, models cannot maintain semantic coherence across multiple subjects.
The second involves multi-reference feature integration: when multiple subjects share the same latent space, their representations naturally interfere, yet existing methods provide no mechanism to explicitly disentangle these conflated features.
\textbf{These observations lead to a critical question: how can we design optimization objectives that simultaneously preserve individual subject fidelity while enforcing inter-subject separability between different subjects' representations?}

To address these fundamental limitations, we propose \textbf{MOSAIC}, a principled framework that reformulates multi-subject personalized generation as a representation optimization problem. 
The foundation of MOSAIC is the establishment of explicit semantic correspondences through a carefully curated dataset featuring densely annotated alignment points across reference-target image pairs. This correspondence foundation enables direct supervision of attention mechanisms for both reference-target alignment and inter-subject differentiation—a critical capability absent in existing approaches.
Leveraging this semantic point correspondence, MOSAIC implements two complementary optimization objectives. The alignment objective employs cross-entropy supervision over attention distributions conditioned on correspondence labels, enforcing precise spatial mappings between reference and target representations. This mechanism ensures semantically coherent feature propagation from reference images to their designated target locations. 
Simultaneously, the disentanglement objective maximizes inter-subject attention divergence via symmetric KL regularization, promoting orthogonal attention patterns across reference subjects. This formulation effectively mitigates cross-subject feature interference—a persistent challenge in multi-reference scenarios.
\textbf{Notably}, our method enables high-quality consistent generation with 4+ reference subjects, a capability that current approaches cannot achieve. Our contributions are as follows:

\begin{itemize}
    \item We propose \textbf{MOSAIC}, a representation-centric framework that learns subject-consistent, disentangled representations by explicitly supervising attention correspondence between target and reference images, combining alignment and disentanglement objectives.
    \item We introduce \textbf{SemAlign-MS}, the first large-scale, comprehensively annotated dataset with fine-grained semantic correspondences specifically tailored for multi-subject driven generation.
    \item We demonstrate significant improvements over existing methods, particularly in challenging scenarios with 4+ subjects, while maintaining computational efficiency through our plug-and-play design.
\end{itemize}
\begin{figure*}[t]
    \centering
    \vspace{-8mm}
    \includegraphics[width=0.95\linewidth]{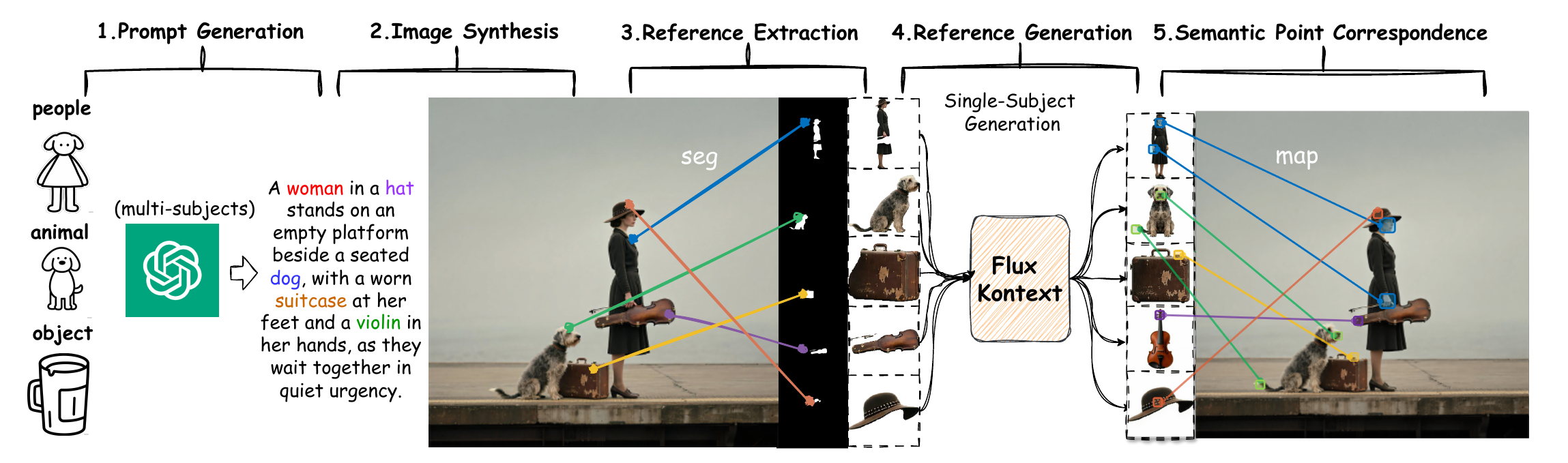}
    \caption{\textbf{SemAlign-MS Dataset Construction Pipeline.} Five-stage systematic pipeline for generating high-quality multi-reference training data with validated semantic correspondences.}
    \label{fig:demo}
\end{figure*}

\section{Related works}

\noindent
\textbf{Subject-Driven Image Generation.}
Subject-driven image generation aims to synthesize images that preserve the identity and appearance of reference subjects while following textual descriptions. Recent advances have focused on improving reference encoding and attention mechanisms within diffusion frameworks. OminiControl~\cite{ominicontrol} exploits the generative model itself as a reference image encoder, demonstrating the capability of diffusion transformers in maintaining subject consistency. For multi-subject scenarios, UNO~\cite{uno} proposes a systematic data generation pipeline, while DreamO~\cite{dreamo} constructs a router mechanism to focus attention on target subjects. XVerse~\cite{xverse} adopts text-stream modulation to transform reference images into token-specific offsets, enabling better integration of subject-specific cues. However, existing methods primarily rely on global feature matching, lacking explicit fine-grained detail constraints between reference and target regions. This limitation often results in imprecise spatial alignment, reduced reference fidelity, and attribute entanglement when handling multiple subjects simultaneously. Our work addresses these fundamental challenges through explicit semantic point correspondences that enable precise spatial alignment and cross-reference disentanglement, substantially enhancing subject consistency while preventing inter-subject interference.

\noindent
\textbf{Visual Correspondence for Generation.}
Visual correspondence establishes spatial relationships between semantically similar regions across images, serving as a foundation for various computer vision tasks~\cite{lee2019sfnetlearningobjectawaresemantic, lee2020referencebasedsketchimagecolorization, lu2024regiondragfastregionbasedimage}. Traditional methods rely on handcrafted features like SIFT~\cite{sift} and SURF~\cite{bay2006surf}, while recent deep learning approaches leverage supervised learning with annotated datasets~\cite{dataset0, dataset1, dataset2}. A promising direction has emerged using diffusion models for correspondence estimation. Methods such as DIFT~\cite{dift}, SD-DINO~\cite{sd-dino}, and GeoAware-SC~\cite{geoaware} demonstrate that pre-trained diffusion features can establish reliable correspondences across diverse images without extensive supervision. However, these correspondences have not been effectively utilized for multi-subject generation tasks. Our work bridges this gap by being the first to leverage semantic point correspondences for multi-subject-driven generation. We establish a systematic pipeline that constructs high-quality correspondences and explicitly incorporates them into the generation process through our semantic corresponding attention alignment and multi-reference disentanglement mechanisms.

\begin{figure*}[t]
    \centering
    \vspace{-8mm}
    \includegraphics[width=0.99\linewidth]{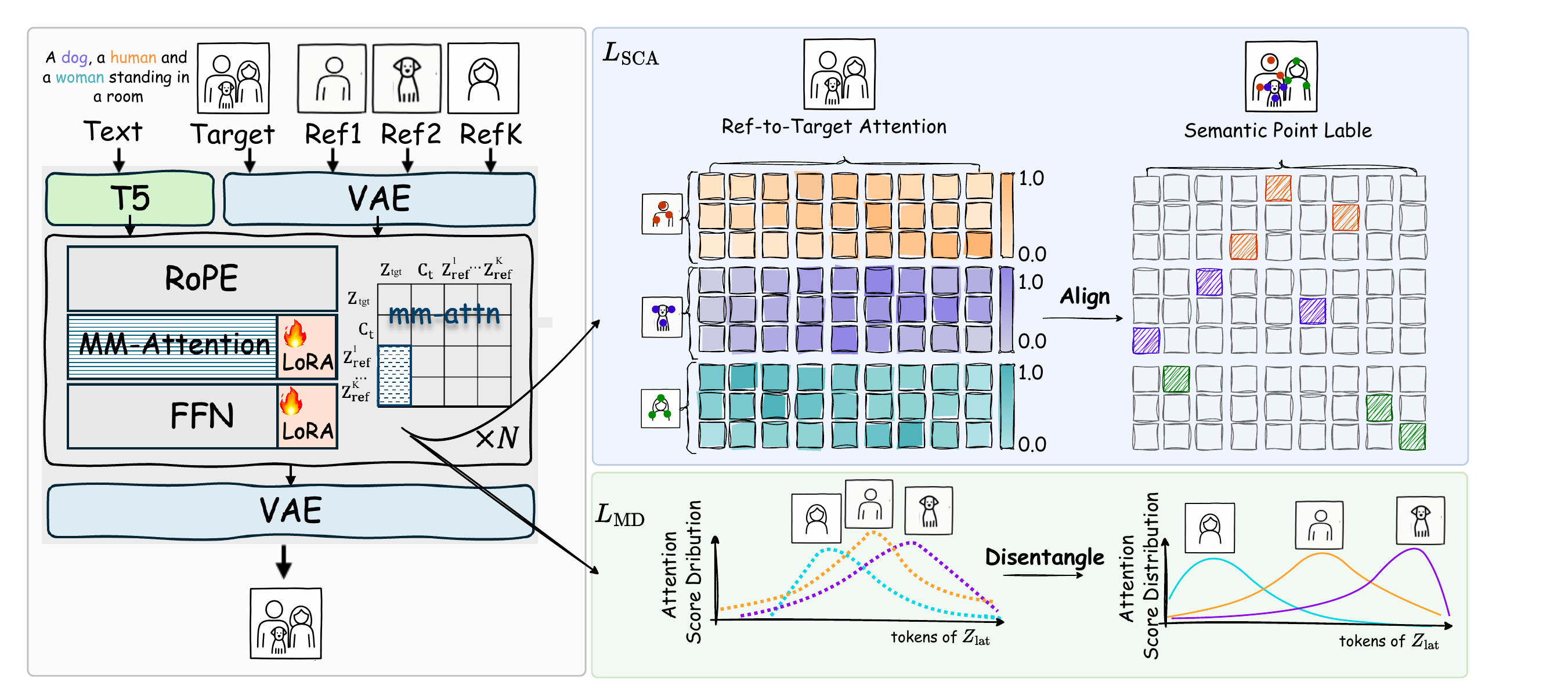}
    \caption{\textbf{Overview of MOSAIC Framework.}  MOSAIC introduces two key supervisions: (1) \textit{Semantic Correspondence Attention Loss} (blue region) enforces precise point-to-point alignment between reference tokens and their corresponding locations in the target latent, ensuring high consistency; (2) \textit{Multi-Reference Disentanglement Loss} (green region) maximizes the divergence between different references' attention distributions, pushing each subject into orthogonal representational subspaces. }
    \label{fig:pipeline}
\end{figure*}

\section{SemAlign-MS: A High-Quality Multi-Subject Dataset with Semantic Point Correspondences}

Our data construction follows a systematic five-stage pipeline designed to ensure both diversity and quality. We first leverage GPT-4o~\cite{gpt4o} with carefully designed templates to automatically generate diverse prompts containing multi-subject, encompassing various combinations of people, animals, objects, and their interactions to ensure comprehensive coverage of multi-subject scenarios commonly encountered in real-world applications. The generated prompts are then processed through state-of-the-art T2I models to synthesize images, where we implement a multi-criteria automated filtering strategy that evaluates image quality, subject clarity, and compositional coherence to retain only the highest-quality synthetic images. Subsequently, we employ Lang-SAM~\cite{SAM} for robust open-vocabulary detection and segmentation of all subjects within the synthesized images, enabling precise identification and isolation of individual subjects regardless of their semantic category and providing the foundation for subsequent correspondence establishment.
Finally, we utilize FLUX Kontext~\cite{flux_kontext} for viewpoint correction while maintaining semantic consistency, significantly enhancing dataset diversity and ensuring comprehensive appearance coverage across different viewpoints and poses.

Once we obtain the large-scale data, we establish \textbf{\textit{semantic point correspondences}} between each target image and multiple reference images, where each reference image contributes $P^{(k)}$ sampled semantic points from the reference space that are mapped to corresponding locations in the target latent space, and $P^{(k)} \in \mathbb{N}$ is the number of semantic correspondence points for the $k$-th reference image. Formally, let $\mathcal{D} = \{( \{\mathcal{I}_{\text{ref}}^{(i,k)}\}_{k=1}^{K}, \mathcal{I}_{\text{tgt}}^{(i)})\}_{i=1}^{N}$ denote our dataset, where $N$ is the total number of target images in the training batch, $M$ is the maximum number of reference images, $\mathcal{I}_{\text{tgt}}^{(i)}$ represents the $i$-th target image, and $\mathcal{I}_{\text{ref}}^{(i,k)}$ denotes the $k$-th reference image for the $i$-th target. When fewer than $K$ reference images are available, we pad the remaining slots with black placeholder images to maintain consistent batch dimensions.  For each pair $( \mathcal{I}_{\text{ref}}^{(i, k)}, \mathcal{I}_{\text{tgt}}^{(i)})$, we define the semantic correspondence set as:
\begin{equation} \label{eq:correspondence}
\mathcal{C}^{(i,k)} = \{(u_{i,j}, v_{i,j})\}_{j=1}^{P^{(k)}}
\end{equation}
where $u_{i,j} $ represents the $j$-th sampled point coordinate in reference image $\mathcal{I}_{\text{ref}}^{(k)}$, and $v_{i,j}$ denotes its corresponding semantic point in the target latent space. To prevent inter-reference conflicts where references map to the same target token, we enforce \textit{\textbf{correspondence disjointness}} across different reference images:
\begin{equation}
\begin{split}
\mathcal{V}^{(i,k)} \triangleq \{v_{i,j}~|~(u_{i,j}, v_{i,j}) \in \mathcal{C}^{(i,k)}\}, \\
\text{s.t.} \quad \mathcal{V}^{(i,k_1)} \cap \mathcal{V}^{(i,k_2)} = \emptyset, \quad \forall k_1 \neq k_2,
\end{split}
\end{equation}
where $\mathcal{V}^{(i,k)}$ represents the  token position corresponding to the $j$-th sampled token of $k$-th reference image for the $i$-th training sample. This constraint ensures that each target latent location $v_{i,j}$ is associated with at most one reference image, preventing ambiguous supervision where multiple references compete for the same target region. Through this systematic pipeline, we successfully collect 1.2M high-quality image pairs with validated semantic correspondences that form the foundation of SemAlign-MS dataset.

\section{Methodology}

\subsection{Overview of MOSAIC}

\par 
\noindent
\textbf{Architecture.}
As shown in Fig.~\ref{fig:pipeline}, given $K$ reference images $\{\mathbf{I}_{\text{ref}}^{(k)}\}_{k=1}^K$ depicting different subjects, a text prompt $\mathbf{T}$, and a target image $\mathbf{I}_{\text{tgt}}$ to be generated. During training, noise is applied to the target image to obtain $\mathbf{I}_{\text{tgt}} = \mathbf{I}_{\text{tgt}} + \epsilon$, where $\epsilon \sim \mathcal{N}(0, 1)$. The VAE encoder then transforms both the noisy image and all reference images into latent representations:
\begin{equation}
\begin{aligned}
\mathbf{z}_{\text{tgt}} = \text{VAE}_{\text{enc}}(\mathbf{I}_{\text{tgt}}),~\mathbf{z}_{\text{ref}}^{(k)} = \text{VAE}_{\text{enc}}(\mathbf{I}_{\text{ref}}^{(k)})
\end{aligned}
\end{equation}
where $ k = 1, \ldots , K$. The text prompt is encoded using T5 \cite{t5} to obtain text embeddings $\mathbf{C}_{\text{t}} = \text{T5}(\mathbf{T})$. To ensure spatial disentanglement between reference and target latents, modified Rotary Position Embeddings (RoPE) \cite{rope} with distinct frequency bases are applied.
\par
\noindent
\textbf{Multi-Reference within MM-Attention.} The core of MOSAIC lies in how multiple reference latent and target latent interact through attention. Following OmniControl \cite{ominicontrol}, we employ a LoRA-augmented branch for reference processing while maintaining the original model weights for the denoising branch. Crucially, we concatenate all reference latents into a unified representation to enable joint processing. In each transformer block $l$, the attention computation proceeds as:

\begin{equation}
\begin{aligned}
    &\mathbf{Q}_{\text{tgt}}^{(l)}, \mathbf{K}_{\text{tgt}}^{(l)}, \mathbf{V}_{\text{tgt}}^{(l)} = f_{\theta}(\mathbf{z}_{\text{tgt}}^{(l)}), \\
    &\mathbf{Q}_{\text{text}}^{(l)}, \mathbf{K}_{\text{text}}^{(l)}, \mathbf{V}_{\text{text}}^{(l)} = f_{\phi}(\mathbf{C}_{\text{t}}),
\end{aligned}
\end{equation}
where $f_{\theta}$ and $f_{\phi}$ mean the mm-attention projection weights for target latent and text embedding.
For the reference images, we first concatenate their latent representations:
\begin{equation}
\mathbf{z}_{\text{ref}}^{(l)} = [\mathbf{z}_{\text{ref}}^{(1,l)}; \mathbf{z}_{\text{ref}}^{(2,l)}; \ldots; \mathbf{z}_{\text{ref}}^{(K,l)}].
\end{equation}
Then process them through the LoRA branch of $f_{\theta}$, which denoted as $\Delta\theta_{\text{LoRA}}$:
\begin{equation}
\mathbf{Q}_{\text{ref}}^{(l)}, \mathbf{K}_{\text{ref}}^{(l)}, \mathbf{V}_{\text{ref}}^{(l)} = f_{\theta + \Delta\theta_{\text{LoRA}}}(\mathbf{z}_{\text{ref}}^{(l)}).
\end{equation}
These features are then concatenated across modalities and processed through multi-head attention:
\begin{equation}
\begin{aligned}
    \mathbf{Q}^{(l)} &= [\mathbf{Q}_{\text{tgt}}^{(l)}; \mathbf{Q}_{\text{text}}^{(l)}; \mathbf{Q}_{\text{ref}}^{(l)}], \\
    \mathbf{K}^{(l)} &= [\mathbf{K}_{\text{tgt}}^{(l)}; \mathbf{K}_{\text{text}}^{(l)}; \mathbf{K}_{\text{ref}}^{(l)}], \\
    \mathbf{V}^{(l)} &= [\mathbf{V}_{\text{tgt}}^{(l)}; \mathbf{V}_{\text{text}}^{(l)}; \mathbf{V}_{\text{ref}}^{(l)}].
\end{aligned}
\end{equation}
The attention scores are computed as $\mathbf{A}^{(l)} = \text{softmax}(\mathbf{Q}^{(l)}(\mathbf{K}^{(l)})^T / \sqrt{d})$, where $d$ is the feature dimension. Note that within $\mathbf{A}^{(l)}$, the reference-to-target latent attention sub-matrix $\mathbf{A}_{\text{ref}\rightarrow\text{tgt}}^{(l)} \in \mathbb{R}^{N_{\text{ref}} \times N_{\text{tgt}}}$ captures how each token from all reference images attends to the noisy latent, where $N_{\text{ref}}$ is the total number of tokens length across all reference images and $N_{\text{tgt}}$ is the token length of the target latent.

\subsection{Semantic Corresponding Attention Alignment}
To faithfully preserve fine-grained semantic details, especially in regions requiring precise structural correspondence, we design the \textbf{\textit{semantic correspondence attention loss (SCAL)}} to explicitly enforce point-wise semantic alignment within the reference-to-target attention mechanism.

As shown in Fig.~\ref{fig:pipeline}, the reference-to-target latent attention allows each token from the reference images to attend to all token positions of the noised target latent.  For a reference token at position $u$ and a target latent position $v$, the average reference to target latent attention across all DiT blocks $N_\text{block}$ is computed as:
\begin{equation}
\mathbf{A}_{\text{ref}\rightarrow\text{tgt}}[u, v] = \frac{1}{N_{\text{block}}} \sum_{l=1}^{N_{\text{block}}}
\frac{\exp\left({\boldsymbol{Q}_u \boldsymbol{K}_v^\top}/{\sqrt{d}}\right)
}{
    \sum_{v=1}^{N_{\text{tgt}}} \exp\left({\boldsymbol{Q}_u \boldsymbol{K}_v^\top}/{\sqrt{d}}\right)
}
\end{equation}
where $\boldsymbol{Q}_u$ and $\boldsymbol{K}_v$ are the query and key embeddings for the reference token at position $u$ and latent position $v$, respectively. To map local reference coordinates to global token indices in concatenated reference representation, we define:
\begin{equation}
\text{G}(u_{i,j}^{(k)}) = \underbrace{\sum_{idx=1}^{k-1} N^{(idx)}}_{\text{offset}} + u_{i,j}^{(k)}
\end{equation}
where $idx$ denotes the reference subject index.
For each correspondence pair $(u_{i,j}^{(k)}, v_{i,j}^{(k)})$, to supervise the attention from the reference token at position $u_{i,j}^{(k)}$ that focused on its corresponding latent position $v_{i,j}^{(k)}$, we define:
\begin{equation}
    \mathcal{L}_{\text{SCA}} = -\frac{1}{K}\sum_{k=1}^{K} \frac{1}{P^{(k)}}\sum_{j=1}^{P^{(k)}} \log \mathbf{A}_{\text{ref}\rightarrow\text{tgt}}[\text{G}(u_{i,j}^{(k)}), v_{i,j}^{(k)}]
\end{equation}
By integrating $\mathcal{L}_{\text{SCA}}$ into the training objective, our model is effectively encouraged to learn precise semantic mappings between reference and generated images.
This leads to significantly improved preservation of local structure, textures, and fine details, going beyond the limitations of global similarity or implicit feature alignment.

\subsection{Multi-Reference Corresponding Disentanglement}
While alignment ensures high consistency, a crucial aspect of multi-subject driven generation is the potential interference among different reference images.
We introduce \textbf{\textit{multi-reference disentanglement loss (MDL)}}, which promotes distinct attention patterns across references. MDL emphasizes the differentiation of attention maps from various subjects, thus preventing feature conflation.

Specifically, for $k$-th reference image, we collect the attention patterns at correspondence locations.
For each correspondence point $(u_j, v_j) \in \mathcal{C}^{(i,k)}$:
\begin{equation}
\boldsymbol{a}^{(k)}_j = [\mathbf{A}_{\text{ref}\rightarrow\text{tgt}}[ u_j, t]~|~t \in N_{tgt}] \in \mathbb{R}^{N_{tgt}}
\end{equation}

Subsequently, we aggregate these attention responses for each reference:
\begin{equation}
\boldsymbol{a}^{(k)} = || \frac{1}{P^{(k)}} \sum_{j=1}^{P^{(k)}} \boldsymbol{a}^{(k)}_j \in \mathbb{R}^{N_{tgt}} ||
\end{equation}
where $|| \cdot ||$ denotes the normalization operation. Then the distance between $a^{(i)}$ and $a^{(j)}$ is:
\begin{equation}
\text{dist}(\boldsymbol{a}^{(i)}, \boldsymbol{a}^{(j)}) = \frac{1}{2}\mathcal{D}_\text{KL}(\hat{\boldsymbol{a}}^{(i)} || \hat{\boldsymbol{a}}^{(j)}) + \frac{1}{2}\mathcal{D}_\text{KL}(\hat{\boldsymbol{a}}^{(j)} || \hat{\boldsymbol{a}}^{(i)})
\end{equation}
where $\mathcal{D}_\text{KL}$ is KL divergence.
Subsequently, we enforce distinct attention patterns across references:
\begin{equation}
\mathcal{L}_\text{MD} = -\frac{1}{K(K-1)} \sum_{i=1}^{K} \sum_{j=1,j \neq i}^{K} \text{dist}(\boldsymbol{a}^{(i)}, \boldsymbol{a}^{(j)})
\end{equation}
By maximizing attention pattern divergence, this loss prevents references from competing for the same attention regions, mitigating cross-reference feature interference.
The overall loss is defined as:
\begin{equation} \label{eq: overall loss}
    \mathcal{L} = \mathcal{L}_\text{diff} + \alpha  \mathcal{L}_\text{SCA} + \beta \mathcal{L}_\text{MD}
\end{equation}
where $\mathcal{L}_\text{diff}$ is the flow-matching loss used in \cite{sd3}, $\alpha$ and $\beta$ are two factors that balance the weight.

\section{Experiments}

\begin{table}[!t]
    \centering
    \setlength{\tabcolsep}{18pt}
    \vspace{-8mm}
        \begin{tabular}{ll|ccc}
        \toprule
        \textbf{Reference} & \textbf{Method} & \textbf{CLIP-I $\uparrow$} & \textbf{CLIP-T $\uparrow$} & \textbf{DINO $\uparrow$} \\
        \midrule
        \multirowcell{9}[0pt][c]{\rotatebox{90}{Single-Subject}}
        & DreamBooth & 80.30 & 30.52 & 66.81 \\
        & BLIP-Diffusion & 80.47 & 30.24 & 69.82 \\
        & SSR-Encoder & 82.10 & 30.79 & 61.22 \\
        & MS-Diffusion & 80.82 & \underline{31.05} & 70.32 \\
        & UNO & \underline{83.50} & 30.41 & 75.97 \\
        & DreamO & 83.35 & 30.61 & \underline{76.03} \\
        & XVerse & 83.20 & 30.20 & 75.44 \\
        & \textbf{MOSAIC} & \textbf{84.30} & \textbf{31.64} & \textbf{77.40} \\
        \midrule
        \multirowcell{5}[0pt][c]{\rotatebox{90}{Multi-Subject}}
        & MS-Diffusion & 72.60 & 31.91 & 52.50 \\
        & UNO & {73.29} & \underline{32.23} & \underline{54.22} \\
        & DreamO & 73.32 & 32.10 & 52.17 \\
        & XVerse & \underline{73.47} & 31.20 & 53.71 \\
        & \textbf{MOSAIC} & \textbf{76.30} & \textbf{32.40} & \textbf{56.83} \\
        \bottomrule
        \end{tabular}
    \caption{Quantitative comparison for single-subject and multi-subject on DreamBench benchmark. }
    \label{tab:quan_cmp_combined}
\end{table}

\begin{table*}[!t]
 \centering
\setlength{\tabcolsep}{4pt}
\begin{tabular}{lccccc|ccccc|c}
\toprule
\multirow{2}{*}{\textbf{Method}}& \multicolumn{5}{c}{\textbf{Single-Subject}}    & \multicolumn{5}{c}{\textbf{Multi-Subject}} & \multirow{2}{*}{\textbf{Overall}}\\
 \cmidrule(lr){2-6} \cmidrule(lr){7-11}
 & DPG & ID-Sim & IP-Sim & AES & AVG & DPG & ID-Sim & IP-Sim & AES & AVG & \\
\midrule
     MS-Diffusion & \underline{96.89} & 6.52 & 55.71 & \underline{59.63} & 54.69 & 87.21 & 3.77 & 46.21 & \textbf{55.91} & 48.28 & 51.49 \\
     UNO  & 89.65 & 47.91& \underline{80.40}  & 55.90 &68.47&85.28&31.82&67.00& 54.24 &59.59 &64.03\\
     DreamO  & \textbf{96.93} & 75.48 & 70.84 & {54.57} & 74.46 & {88.80} & 50.24 & {64.63} & 52.47 & {64.04} & 69.25 \\
     OmniGen2 & 92.60 & 62.41 & 74.08&52.34& 70.36 & \textbf{91.55}&40.81& 67.15 &51.40&62.73 & 66.55\\
     XVerse & 93.69 & \underline{79.48} & 76.86 & 56.84 & \underline{76.72} & 88.26 & \underline{66.59} & \underline{71.48} & 53.97 & {70.08} & \underline{73.40} \\
\midrule
    \textbf{MOSAIC} & 96.55 & \textbf{81.98} & \textbf{{80.92}} & \textbf{60.77} & \textbf{80.05} & \underline{88.94} & \textbf{69.90} & \textbf{74.27} & \underline{55.02} & \textbf{72.03} & \textbf{76.04} \\
 \bottomrule
\end{tabular}
 \caption{Quantitative results of single-subject and multi-subject driven generation on XVerseBench.}
 \label{tab:combined_single_multi}
 \end{table*}

\subsection{Experimental Details}

\begin{figure*}[!t]
    \centering
    \vspace{-8mm}
    \includegraphics[width=0.99\linewidth]{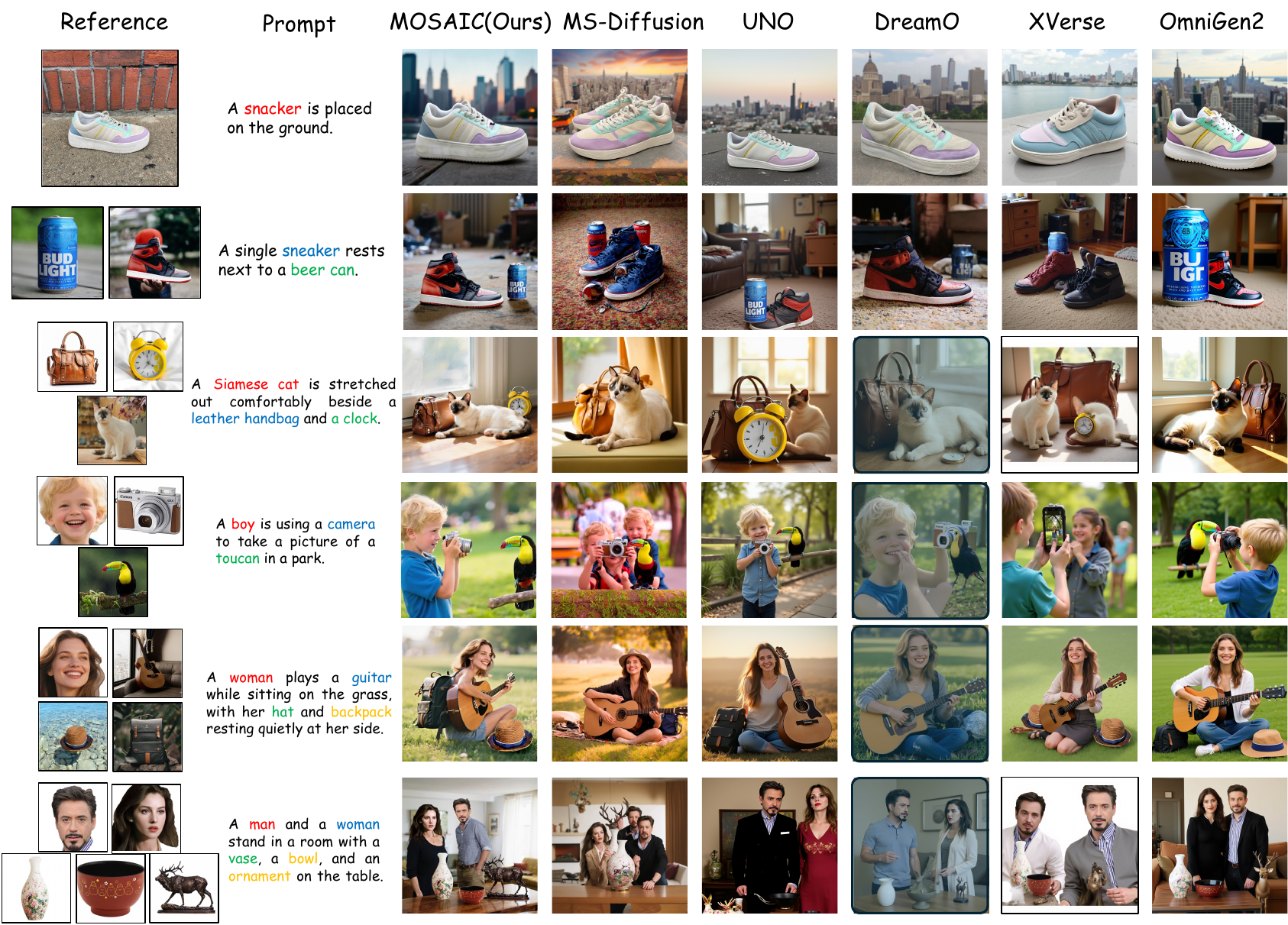}
    \caption{Qualitative comparison on single and multi-subject driven generation. DreamO does not support 3+ reference images; masked regions show where comparison is unavailable, with visible results generated from two randomly selected references.}
    \label{fig: main_fig}
\end{figure*}

\subsubsection{Implementation Details}
In this work, we adopt FLUX-1.0-DEV~\cite{flux} as our base model. The rank of additional LoRA~\cite{lora} is set as 128.
During training, we employ the AdamW~\cite{adam} optimizer with a learning rate of 1e-4 and train the model for 100K steps, using a batch size of 1 per GPU.
The $\alpha$ and $\beta$ in Eq.~\ref{eq: overall loss} are set to 0.4 and 0.6.

\subsubsection{Evaluation Setting}
We evaluate our proposed method on DreamBench~\cite{dreambooth} and XVerseBench~\cite{xverse}, focusing on both single-subject and multi-subject generation scenarios.
For comparison, we include DreamBooth~\cite{dreambooth}, BLIP-Diffusion~\cite{blip-diffusion}, SSR-Encoder~\cite{ssr-encoder}, MS-Diffusion~\cite{wang2024ms}, UNO~\cite{uno}, DreamO~\cite{dreamo}, OmniGen2~\cite{omniGen2}, and XVerse~\cite{xverse}.
DreamBench utilizes the CLIP-I~\cite{CLIP}, DINO~\cite{dino}, and CLIP-T~\cite{CLIP} to evaluate the semantic and text-image alignment.
XVerseBench utilizes DPG score~\cite{ella}, ID-Sim~\cite{deng2019arcface}, IP-Sim~\cite{dino}, AES~\cite{discus0434_aesthetic_2024} to further evaluate the alignment and image quality.

\subsection{Main Results}

\subsubsection{Quantitative Results}
The results on DreamBench and XVerseBench are shown in Tab.\ref{tab:quan_cmp_combined} and Tab.\ref{tab:combined_single_multi}, respectively.
On DreamBench, MOSAIC achieves consistently strong results across all metrics. In the single-subject setting, it achieves 84.30 (CLIP-I), 31.64 (CLIP-T), and 77.40 (DINO), surpassing the second-best method UNO by 0.80 on CLIP-I, and the second-best method DreamO by 1.37 on DINO, indicating better image-text alignment and semantic fidelity.
In the more challenging multi-subject setting, it maintains strong performance with 76.30 (CLIP-I), 32.40 (CLIP-T), and 56.83 (DINO), outperforming second-best competitors by approximately 3.0, 0.2, and 2.6 points, respectively.
On XVerseBench, MOSAIC achieves the highest overall average score of 76.04, surpassing XVerse's 73.40. And MOSAIC achieves a DPG score that is on par with the top-performing DreamO (96.93). It also demonstrates clear advantages in identity preservation, with ID-Sim scores of 81.98 for single-object and 69.90 for multi-object scenarios, as well as in perceptual similarity, with IP-Sim scores of 80.92 and 74.27, respectively.
Overall, these demonstrate the robustness of our proposed method under varying subject complexity.

\subsubsection{Qualitative Results}
Fig.~\ref{fig: main_fig} presents qualitative results across scenes with varying numbers of reference objects, demonstrating MOSAIC's superior performance in multi-subject generation.
\textbf{Appearance consistency.} Our method maintains strong visual coherence across different subjects. For instance, the snackers in rows 1-2 and the can in row 2 preserve consistent textures and shapes throughout the generated scenes, indicating effective feature transfer from reference images.
\textbf{Multi-subject handling.} The advantages of our approach become particularly evident with three or more reference subjects. In row 3, competing methods suffer from object omission and duplication: MS-Diffusion and OmniGen2 fail to render the clock entirely, while XVerse produces duplicated and deformed cats. In contrast, MOSAIC accurately represents all three objects without such artifacts.
\textbf{Scalability to more complex scenes.} For scenes involving four or more reference objects, existing methods exhibit substantial degradation. As shown in row 6, only MOSAIC preserves facial consistency across multiple subjects, while other approaches show significant quality loss and identity confusion.
Overall, these results highlight MOSAIC's ability to generate high-quality, consistent outputs across diverse compositional scenarios, especially in challenging multi-subject settings where other methods fall short.

\subsection{Ablation Studies}

\begin{table}[!t]
\centering
\vspace{-8mm}
\setlength{\tabcolsep}{20pt}
\begin{tabular}{cc|ccc}
\toprule
$\mathcal{L}_{\text{SCA}}$ & $\mathcal{L}_{\text{MD}}$  & \textbf{CLIP-I} & \textbf{CLIP-T} & \textbf{DINO} \\
\midrule
\xmark & \xmark  & 73.45 & 29.90 & 52.03 \\ 
\cmark & \xmark  & 75.89
& 31.10 & 55.99 \\ 
\cmark & \cmark & 76.30 & 32.40 & 56.83  \\ 
\bottomrule
\end{tabular}
\caption{Impacts of different losses on DreamBench benchmark in multi-subject scenario.}
\label{tab:ablation1}
\end{table}



\begin{figure}[t]
    \centering
    \begin{minipage}{0.40\textwidth}
        \centering
        \includegraphics[width=0.99\linewidth]{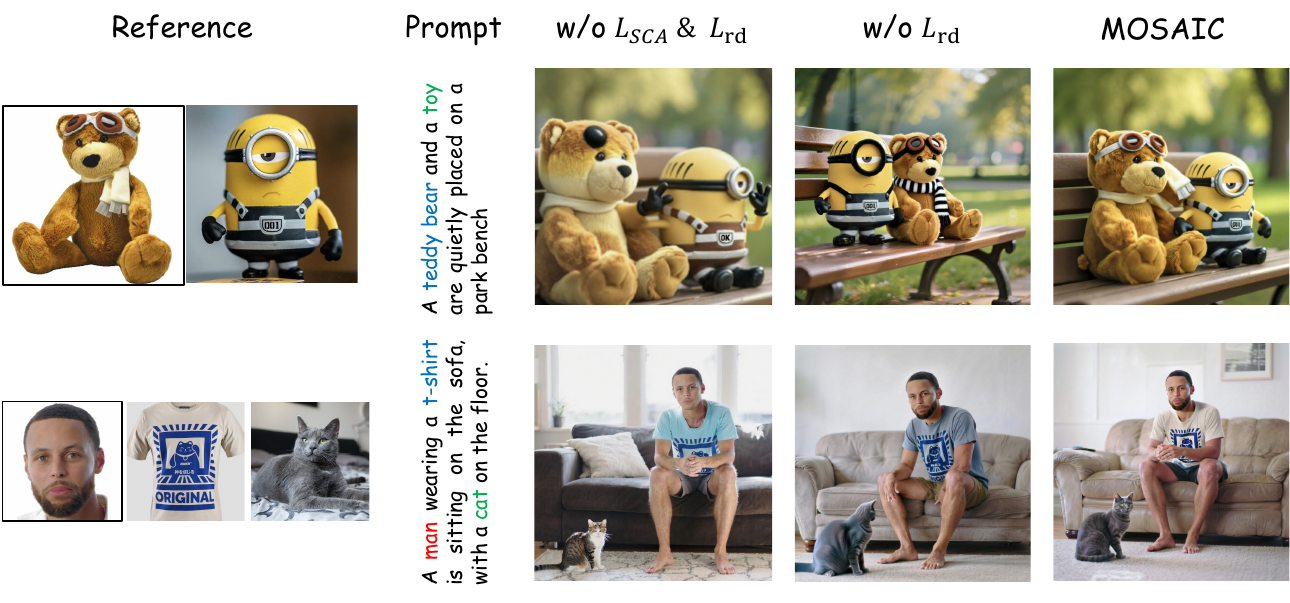}
        \caption{Ablation study of MOSAIC. Zoom in for details.}
        \label{fig:ablation}
    \end{minipage}
    \hfill
    \begin{minipage}{0.58\textwidth}
        \centering
        \includegraphics[width=0.99\linewidth]{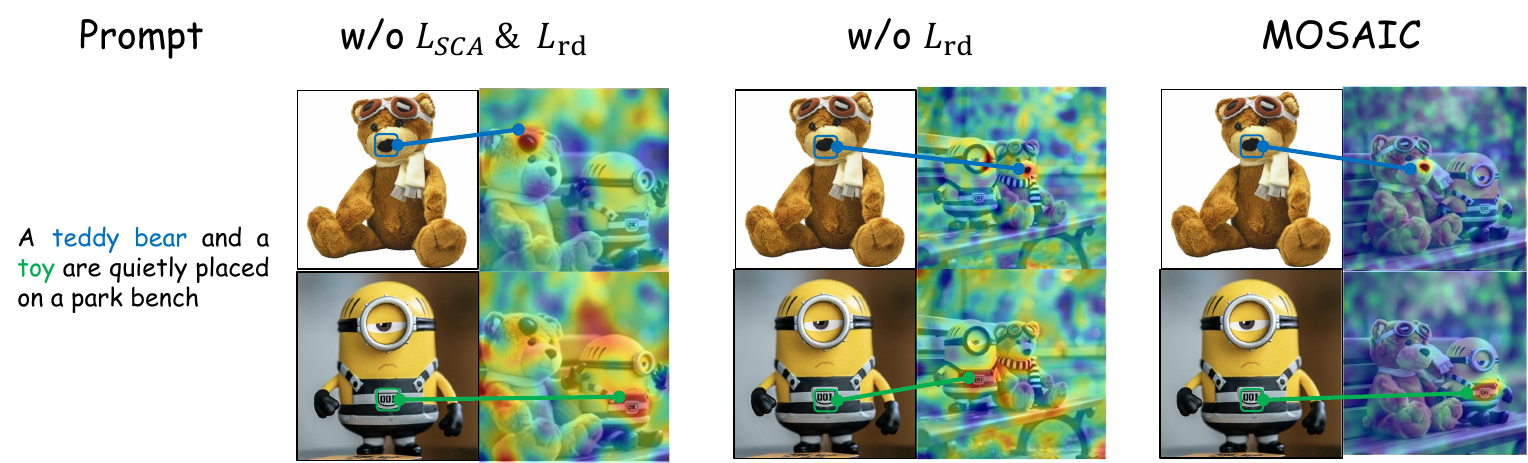}
        \caption{Visualization of attention maps from specific reference regions to generated image regions.}
        \label{fig:ablation2}
    \end{minipage}
\end{figure}

\subsubsection{Impact of Different Losses}
We conduct ablation studies to evaluate each component of our method on multi-subjects of Dreambench, as shown in Tab.\ref{tab:ablation1}.
Adding the semantic correspondence attention loss $\mathcal{L}_{SCA}$ leads to consistent gains across all metrics, with CLIP-I improving from 73.45 to 75.89, CLIP-T from 29.90 to 31.10, and DINO from 52.03 to 55.99. This confirms that explicit point-wise semantic supervision aids fine-grained detail retention. Moreover, introducing  $\mathcal{L}_{MD}$ further improves CLIP-I to 76.30, CLIP-T to 32.40, and DINO to 56.83, suggesting that suppressing cross-reference feature conflation strengthens semantic fidelity and compositional coherence.
Moreover, visual results in Fig.~\ref{fig:ablation} improve progressively: the baseline yields blurry, inconsistent outputs, while adding $\mathcal{L}_{\text{SCA}}$ enhances composition and detail. The full MOSAIC model achieves both global scene coherence and local identity preservation by preventing semantic interference across references.

\subsubsection{Visualization of attention maps}
As shown in Fig.~\ref{fig:ablation2}, progressive improvement in attention alignment and disentanglement through our proposed losses. We visualize attention maps from specific reference regions (teddy bear's goggles and Minion's "001" text) to demonstrate the effectiveness of our semantic corresponding attention loss ($\mathcal{L}_\text{SCA}$) and multi-reference disentanglement loss ($\mathcal{L}_\text{rd}$). From left to right: baseline method shows scattered, unfocused attention with cross-reference interference; $\mathcal{L}_\text{SCA}$ improves semantic alignment but attention remains diffuse across multiple regions; $\mathcal{L}_\text{SCA}$ + $\mathcal{L}_\text{rd}$ achieves both precise attention alignment to semantically corresponding regions and effective disentanglement between different reference subjects. Blue and green lines connect reference regions to their attention peaks in the target latent space.
\section{Conclusion}

In this paper, we propose MOSAIC, a representation-centric approach for multi-subject driven image generation.
Our method features a semantic correspondence pipeline, along with explicit attention alignment and disentanglement mechanisms, effectively addressing the key challenges of identity preservation and attribute entanglement in multi-subject personalized generation scenarios.
We also curate and will release SemAlign-MS, a large-scale multi-subject dataset with fine-grained semantic point correspondences to facilitate future research in controllable generation. Extensive experiments demonstrate that MOSAIC achieves state-of-the-art performance in both identity fidelity and semantic consistency across established benchmarks. Notably, our approach exhibits robust scalability, successfully handling complex compositions with \textbf{4+} reference subjects.


\bibliographystyle{plainnat}
\bibliography{main}




\end{document}